\documentclass{article}
\usepackage{amsmath}
 \usepackage[main, final]{neurips_2026}
\usepackage[utf8]{inputenc} % allow utf-8 input
\usepackage[T1]{fontenc}    % use 8-bit T1 fonts
\usepackage{hyperref}       % hyperlinks
\usepackage{url}            % simple URL typesetting
\usepackage{booktabs}       % professional-quality tables
\usepackage{amsfonts}       % blackboard math symbols
\usepackage{nicefrac}       % compact symbols for 1/2, etc.
\usepackage{microtype}      % microtypography
\usepackage{xcolor}         % colors
\usepackage{graphicx} 
\usepackage{bm} 
\usepackage{multirow}
\usepackage{array}
\usepackage{hhline}
\usepackage{xcolor}

\usepackage{adjustbox}
\title{LLM Compression by Block Removal with Constrained Binary Optimization}
\author{%
  David Jansen  \\
  Multiverse Computing\\
  \texttt{david.jansen@multiversecomputing.com} \\
   \And
  Roman Rausch  \\
  Multiverse Computing\\
     \And
     Ali Hashemi   \\
     Multiverse Computing\\
     \And
    David Montero  \\
    Multiverse Computing\\
     \And
Rom{\'a}n Or{\'u}s  \\
Multiverse Computing\\
Donostia International Physics Center \\
Ikerbasque Foundation for Science  \\
}

\begin{document}

\maketitle

\begin{abstract}
In this paper, we formulate the compression of large language models (LLMs) by optimally deleting transformer blocks (``block removal'') as a constrained binary optimization (CBO) problem that can be mapped to a physical system (Ising glass), whose energies are a strong proxy for downstream model performance. %2 method introduction
This formulation enables an efficient ranking of a large number of candidate block-removal configurations yielding many high-quality, non-trivial solutions beyond those only removing consecutive regions. %3 method in more detail
Our method performs strongly in the deep compression regime, such as for 50\% compression of Llama-3.3-70B-Instruct, where we achieve an almost 23 percentage point increase on the MMLU benchmark compared to other state-of-the-art (SOTA) block-removal methods.
For lighter compression, it performs on par with those methods across several benchmarks for Llama-3.1-8B-Instruct, Qwen3-14B (both before and after retraining), as well as Llama-3.3-70B-Instruct.  %4 results
The approach is computationally efficient and requires only forward and backward passes on a calibration dataset for a few active parameters. Additionally, we demonstrate that using good heuristic solvers for the CBO problem provides solutions that perform well on downstream tasks in negligible runtime when it is unfeasible to solve the problem exactly. %5 cost
The method can be readily applied to any architecture. We illustrate this generality on the recent NVIDIA-Nemotron-3-Nano-30B-A3B-FP8 model, which exhibits a highly inhomogeneous and challenging block structure, and where we outperform SOTA for AIME25 and GPQA when removing either 2 attention layers or 3 mixture-of-experts layers. %6 additional result
\end{abstract}

\section{Introduction}

   %Large language models (LLMs) have been one of the most disruptive technological advances of this decade. Built primarily on the transformer architecture~\cite{vaswani2017attention}, they have demonstrated impressive capabilities in writing, summarization, and problem-solving~\cite{openai_23}. Over the last few years, the number of released models has grown rapidly~\cite{meta_llama_24,qwen3_25,openai_25}. However, frontier models usually contain billions or even trillions of parameters, making them expensive to train and deploy, which greatly limits their use on resource-constrained devices. Consequently, substantial research effort has focused on compressing LLMs while preserving performance.
Large language models (LLMs) have demonstrated impressive capabilities in writing, summarization, and problem-solving~\cite{openai_23}, and the last few years, the number of released models has grown rapidly~\cite{meta_llama_24,qwen3_25,openai_25}.   However, these frontier models, built primarily on the transformer architecture~\cite{vaswani2017attention}, usually contain billions or even trillions of parameters, making them expensive to train and deploy, which greatly limits their use on resource-constrained devices. Consequently, substantial research effort has focused on compressing LLMs while preserving performance.
   
    LLM compression methods take different approaches to reduce the model size, such as pruning~\cite{sun2023wanda,muralidharan_24,guo2025slimllm,Gili_25}, factorization~\cite{yuan_23,wang2025svdllm,rausch_25}, and block removal~\cite{Ding_25,men2025shortgpt,mistral3_26}. The latter consists of removing whole transformer blocks and is a particularly attractive compression method because, apart from memory savings, it also achieves predictable substantial gains in inference speed due to a shorter model and is compatible with the other compression techniques. However, as models grow deeper and become more heterogeneous, block pruning becomes challenging: the effect of removing a given block depends strongly on which other blocks are removed, which makes the task a hard combinatorial problem.
    
    In this work, we introduce a novel algorithm for selecting combinations of transformer blocks to remove. We associate a binary block-selection variable with each block (0: keep, 1: remove), construct a second-order approximation of the loss by Taylor-expanding it with respect to these variables and compute an approximate Hessian. This maps the original task to a constrained binary optimization (CBO) problem. Although solving the CBO is exponentially complex in the number of blocks, it is substantially cheaper to evaluate a certain configuration (computing its energy) than, for example,  benchmarking the full model. Furthermore, as demonstrated in Sec.~\ref{sec:cbo_solution}, heuristic solvers can often find the optimal (or close to optimal) solution in realistic scenarios within seconds. Crucially, we find that not only the minimum-energy solution, but also a range of low-energy solutions yield high-quality pruning configurations.
    
    Across several benchmarks, these solutions are on par or outperform SOTA block-removal methods~\cite{Ding_25,men2025shortgpt,mistral3_26}. We discuss Llama-3.1-8B-Instruct~\cite{meta_llama_24}, Qwen3-14B~\cite{qwen3_25} with and without light retraining, as well as Llama-3.3-70B-Instruct (without retraining due to its large size). For Llama-3.3-70B-Instruct, we find that our method achieves significantly higher scores across several benchmarks at 50\% compression. This includes a gain of almost 23\% in the MMLU score~\cite{mmlu_20}. 
    
    We finally show that our method easily generalizes beyond dense transformers to heterogeneous architectures, including mixture-of-experts models (MoE). Applying our approach to NVIDIA-Nemotron-3-Nano-30B-A3B-FP8~\cite{nvidia_nemotron_nano_v3_2025} and when comparing to using the block influence method from Ref.~\cite{men2025shortgpt}, we find that our method provides high-quality models with scores higher in both GPQA~\cite{gpqa_23} and AIME25 when either removing 3 mixture-of-expert layers or 2 attention layers.
    
    In summary, our main contributions are:
    \begin{itemize}
    \item We formulate transformer block pruning as a CBO problem by explicitly modelling second-order interactions between blocks, with tractable computational cost for models of practical depth.
    \item We show that the energy of the CBO problem is a strong proxy of the quality of the block-pruned model, in the sense that it correlates well with benchmark scores.
    \item We demonstrate that this formulation outperforms state-of-the-art block-removal techniques across several benchmarks and achieves significantly higher scores for deep compression of the Llama-3.3-70B-Instruct model.
    \item We show that approximate solvers also deliver high-quality candidate solutions, eliminating the need for an exact CBO solver for large models.
    \item We demonstrate that our approach generalizes to heterogeneous MoE architectures, yielding strong performance for NVIDIA-Nemotron-3-Nano-30B-A3B-FP8 and outperforming SOTA without retraining.
    \end{itemize}

% \begin{figure}
%   \centering
%   \includegraphics[width=0.8\textwidth]{work_flow_diagram.pdf}
%   \caption{Sketch of our method. We formulate block pruning as a constrained binary optimization problem that can be mapped to an Ising model with fixed magnetization. Each feasible solution specifies which $M$ out of $N$ blocks are removed from the LLM. The optimization is constructed such that low-energy solutions correspond to pruned models with high performance across multiple benchmarks.}
%   	\label{workflow}
% \end{figure}

%        \begin{figure}
%   \centering
%   \includegraphics[width=\textwidth]{extremlyeasyqubo.pdf}
%   \caption{Illustration of the modification we make to the forward pass of the LLM.}
%   	\label{sketch}
% \end{figure}  

\begin{figure}
  \centering
  \includegraphics[width=\textwidth]{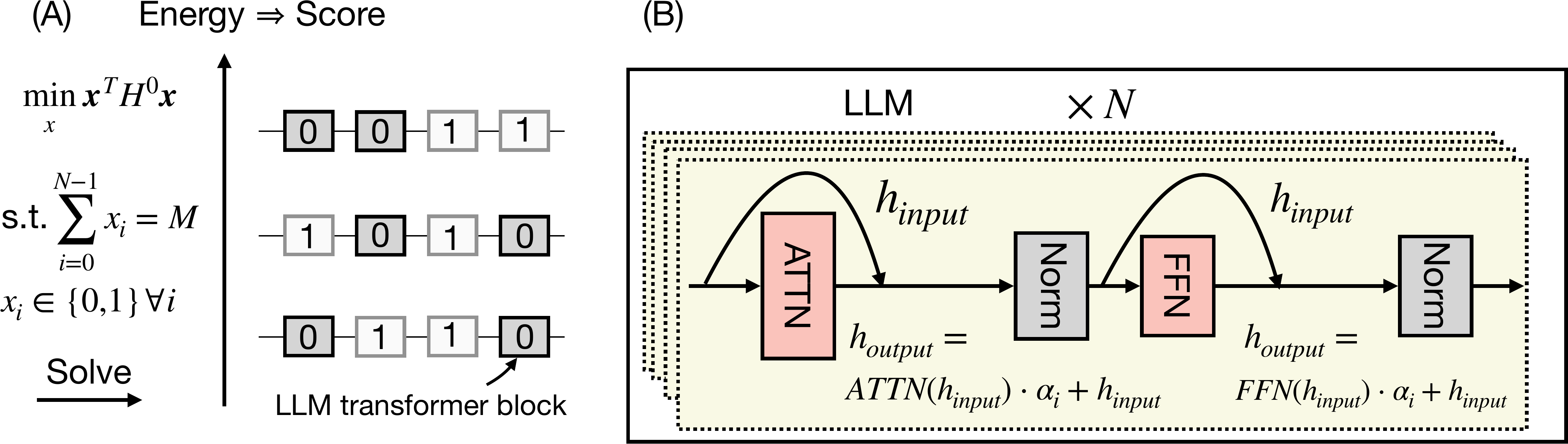}
  \caption{\label{sketch}
  (A) Sketch of our method. We formulate block pruning as a constrained binary optimization (CBO) problem. Each feasible solution specifies which $M$ out of $N$ blocks are removed from the LLM. The optimization is constructed such that low-energy solutions correspond to pruned models with high performance across multiple benchmarks. (B) Illustration of the modification we make to the forward pass of the LLM described in Sec.~\ref{sec:meth}.}
  	
\end{figure}

\section{Related work}
    \label{sec:relwork}
     Existing LLM compression approaches can broadly be categorized into pruning methods, factorization algorithms, accuracy-reduction methods, and structural approaches that remove entire transformer blocks (``depth pruning''). The method presented in this work falls into the last category.
    
    In pruning, one can distinguish between structured (``width pruning'') and unstructured pruning. Whereas the former removes complete rows and columns from the weight matrices, the latter focuses on removing individual weights. In both cases, the decisions are based on magnitude or sensitivity metrics~\cite{sun2023wanda,muralidharan_24,guo2025slimllm,Gili_25}. Factorization algorithms, on the other hand, are often based on the singular value decompositions, approximating weight matrices as the products of two lower-rank matrices~\cite{yuan_23,wang2025svdllm,rausch_25}. Lastly, quantization~\cite{frantar2022gptq,lin2023awq} reduces the numerical precision of the weights and decreases its size without changing the number of parameters and can be combined with the other techniques. 
    %Unstructured pruning methods remove individual weights. Structured pruning methods remove entire channels, often the rows and columns of the weight matrices (``width pruning''). The decision of what to remove is based on magnitude or sensitivity metrics~\cite{sun2023wanda,muralidharan_24,guo2025slimllm,Gili_25}. Low-rank factorization is usually based on the singular value decomposition split the weight matrix into a product of smaller weights~\cite{yuan_23,wang2025svdllm,rausch_25}. Reducing the numerical precision of the weights speeds up the model and decreases its size without changing the number of parameters. The prominent approach here is 
    %quantization~\cite{frantar2022gptq,lin2023awq}, or the recent weight clustering~\cite{aizpurua2026}. Clearly, accuracy reduction techniques can be combined with any weight reduction techniques.
    
    Removing entire transformer blocks directly reduces the model depth and inference cost. Prior work in this area includes analyzing how input representations evolve across blocks~\cite{men2025shortgpt,gromov2025the,mistral3_26}, estimating block importance via various heuristic or sensitivity-based metrics~\cite{kim2024mefomo,song2024sleb}, and incorporating block merging strategies~\cite{Ding_25}. While often effective, these methods generally rely on local or greedy criteria. They do not explicitly take interactions between blocks into account when selecting a global subset to remove, and might therefore be called ``mean-field methods'' in physical terms. Another approach is to restrict the combinations to consecutive blocks only~\cite{gromov2025the}, which goes beyond mean field, but greatly reduces the combinatorial space. As will be shown below, we find that high-quality solutions do not exhibit one monolithic consecutive region that can be removed.
    
    Our approach is inspired by second-order pruning techniques that rely on a Taylor expansion of the loss function. Such methods were first introduced in the context of weight pruning through a technique called \textit{Optimal Brain Damage}~\cite{lecun_89}, and were later extended to combinatorial formulations, such as the \textit{Combinatorial Brain Surgeon}~\cite{Yu_22}, with further scalability improvements via iterative schemes~\cite{Gili_25}. However, these works focus on removing individual weights or channels rather than entire transformer blocks. In contrast, we introduce auxiliary variables that govern the selection of whole blocks. The second-order term of the loss function is then used to explicitly consider interactions between blocks, which goes beyond the previous mean-field approaches and respects the full combinatorial nature of the problem at the same time.

\section{\label{sec:meth}Method}

    Our method extends the ideas from the \textit{Combinatorial Brain Surgeon}~\cite{Yu_22} to block removal. In that work, the authors describe how to identify which weights to prune by solving a combinatorial optimization problem. To adapt this approach to block-level pruning, we introduce a coupling variable $\alpha_i$ for each transformer block $i$. The residual connections in the forward pass of the attention block $ATTN$ and the feed-forward neural network block $FNN$ are modified as follows:
    \begin{equation}
    h_{\text{output}} = ATTN(h_{\text{input}})\cdot \alpha_i + h_{\text{input}} \, ,
    \end{equation}
    \begin{equation}
    h_{\text{output}} = FFN(h_{\text{input}})\cdot \alpha_i + h_{\text{input}} \, .
    \end{equation}
    This modification is illustrated in Fig.~\ref{sketch} (B). We initialize $\alpha_i = 1$ for all $i$, which recovers the original model.
    
    Next, we pass a calibration dataset batch through the model and consider the change in the cost function $\mathcal{L}(\boldsymbol{\alpha})$ (in our case always the cross-entropy) under perturbations $\delta \boldsymbol{\alpha} = \boldsymbol{\alpha} - \boldsymbol{\alpha}^0$ by performing a Taylor expansion~\cite{lecun_89,hassibi_93,kurtic-etal-2022-optimal,Gili_25}:
 
    \begin{equation}
    \delta \mathcal{L}(\boldsymbol{\alpha}) = \mathcal{L}(\boldsymbol{\alpha}) - \mathcal{L}(\boldsymbol{\alpha}^0) \simeq - \beta \delta \boldsymbol{\alpha}^\top \nabla \mathcal{L}(\boldsymbol{\alpha}^0) + \frac{1}{2} \delta \boldsymbol{\alpha}^\top H^0 \delta \boldsymbol{\alpha} + \mathcal{O}(\delta \boldsymbol{\alpha}^3) \, ,
    \label{eq:taylorexp}
    \end{equation}
    where $H^0 = \nabla^2 \mathcal{L}(\boldsymbol{\alpha}^0)$ is the Hessian matrix, and $\beta$ is a hyperparameter. Strict Taylor expansion requires $\beta=1$, but $\beta$ can also be empirically tuned~\cite{Gili_25}, which essentially tunes the relative strength of the first- and second-order terms.
    In our experiments, we find that a finite $\beta$ worsens the results (see App.~\ref{sec:grad} for more details), and we set $\beta=0$ (which was also done in the original work~\cite{lecun_89} based on the assumption that the model is well-trained at a minimum of the loss function). This also simplifies the method, as it eliminates a hyperparameter. We note that this is entirely empirical, however, and that for other problems, tuning $\beta$ might be of benefit.
    
    Similar to Ref.~\cite{Yu_22}, we introduce binary variables $x_i$ with the meaning $x_i = 0$: block is retained, $x_i=1$: block is removed, and write
    \begin{equation}
    \delta \alpha_i = \alpha_i - \alpha_i^0 = (1 - x_i)\alpha_i^0 - \alpha_i^0 = -x_i \alpha_i^0 \, .
    \end{equation}
    We are then left with the following constrained binary optimization problem:
    \begin{equation}
    \label{eq:cbo}
    % \begin{aligned}
    \min_{\boldsymbol{x} \in \{0,1\}^N} %\quad &
    \boldsymbol{x}^\top H^0 \boldsymbol{x} 
    \quad \text{subject to} \quad %&
    \sum_{i=0}^{N-1} x_i = M \, .
    % \end{aligned}
    \end{equation}
    Here, $M$ denotes the targeted number of blocks to be removed, and $N$ the total number of blocks in the model. We further approximate the $(N\times N)$ Hessian as~\cite{spall_05}
    \begin{equation}
    H^0 \simeq \frac{1}{m} \mathcal{A}^\top \mathcal{A} \, ,
    \label{eq:Hessian_approx}
    \end{equation}
    where $\mathcal{A}$ is a matrix of per-sample gradients, and $m$ is the number of samples.
    
    Lastly, we note that that the Hessian only needs to be computed once, after which Eq.~\eqref{eq:cbo} can be solved for a variety of different values of $M$, and with various methods.
    
    \section{\label{sec:results}Experiments}

    We remove 8/32 and 16/32  blocks from Llama-3.1-8B-Instruct~\cite{meta_llama_24}, 12/40 blocks from Qwen3-14B~\cite{qwen3_25}, as well as 8/80, 16/80, 24/80, 32/80 and 40/80 blocks from Llama-3.3-70B-Instruct~\cite{meta_llama_24}. As a calibration dataset for generating the Hessian, we use 2048 samples randomly drawn from OpenHermes-2.5~\cite{OpenHermes25} (we study the effect of different datasets in App.~\ref{sec:dataset}).

    Note that all compression methods lead to a degradation of the model, a large part of which can be mitigated by light retraining on a small computational budget. In particular for deep compression, benchmark scores eventually become random, making a comparison between techniques impossible, but they can quickly recover after a short retraining.
    
    Therefore, we retrain the smaller models using knowledge distillation with the respective uncompressed model as the teacher~\cite{muralidharan_24} for one epoch on OpenHermes-2.5. The teacher models are loaded with 4-bit quantization using the NF4 scheme~\cite{dettmers_23}. We use a global batch size of 64 for Llama-3.1-8B-Instruct and 16 for Qwen3-14B. The remaining hyperparameters are found in our code and in App.~\ref{ret}. We used 4 H200 GPUs for the retraining, and the running time was about 1 day. We do not retrain Llama-3.3-70B-Instruct because of the high computational cost. 
    
    All benchmarks are evaluated using the lm-eval-harness~\cite{eval-harness}. We evaluate MMLU~\cite{mmlu_20},  HellaSwag~\cite{hellaswag_19}, Winogrande~\cite{winogrande_19}, the ARC Challenge~\cite{arc_18}, BBH~\cite{bbh_22} and GSM8K~\cite{gsm8k_21}.

    The full workflow of the method is:
    \begin{enumerate}
    \item Forward pass of the calibration batch through the model with gradient computation that leads to the construction of the Hessian $H^0$ (Eq.~\eqref{eq:taylorexp}) from the gradient (Eq.~\eqref{eq:Hessian_approx}).
    \item Solving the CBO problem Eq.~\eqref{eq:cbo} (minutes to hours for the exact brute-force approach, seconds using the tabu algorithm), see Sec.~\ref{sec:cbo_solution} for more details.
    \item Select candidate solutions (see Sec.~\ref{sec:exc} for more details), remove the blocks and benchmark.
    \item (Optional) Light retraining followed by benchmarking: This is in fact the most costly step, but it is common to all compression methods.
    \end{enumerate}

     We compare our method against block influence (BI)~\cite{men2025shortgpt}, sliding window merging (SWM)~\cite{Ding_25}, and norm ratio~\cite{mistral3_26}, using the same calibration data and retraining procedure. 
     %Block influence is applied iteratively: one block is removed at a time, and the importance scores are recomputed after each removal.

     In App.~\ref{app}, we provide the specific block indices removed by each algorithm.

    \subsection{\label{sec:cbo_solution}Solving the CBO problem}

  Checking each CBO combination just involves the computation of the energy $E=\boldsymbol{x}^\top H^0 \boldsymbol{x}$, where $H^0$ is an $N\times N$ matrix. This is very cheap, but the number of combinations is given by the binomial coefficient $\binom{N}{M}$, which grows exponentially. Luckily, for the block removal problem, $N$ is never exceedingly large (for the models here: $N=32\ldots 80$). Note that the problem is also tractable for large $N$ as long as $M$ is small (i.e. for removing a few blocks from a very deep model).
  
  In our code, we perform a batched matrix-matrix multiplication on an H200 GPU to accelerate the brute-force computation, and are able to fully check combinations up to the order of $O(10^{10})$. We note that the problem is also embarrassingly parallel and could be split over multiple GPUs for additional acceleration, but we did not attempt this.
  
  The computation time is just a couple seconds for a few million combinations, and of the order of minutes below 1B combinations. For the case of 5.59B combinations, it took \textasciitilde3.5 hours. The hardest still tractable problem is the removal of 8/80 blocks of Llama-3.3-70B-Instruct (29B combination), which took about \textasciitilde 2 days. Removing more blocks from Llama-3.3-70B-Instruct becomes intractable.
  
    When dealing with the intractable regime, the problem can be reformulated as a quadratic unconstrained binary optimization (QUBO) problem, where the constraint is now accounted for by a penalty term. In condensed matter physics, this is equivalent to an Ising glass model (all-to-all coupling) with conserved magnetization. This allows us to employ highly optimized classical or quantum solvers~\cite{Farhi_14,Quinton_25,gurobi,liang_libtsqubo,Altelarrea_Ferre_25}. Gurobi~\cite{gurobi} is commonly traded as the best solver on the market, and can be used without license for up to $N=200$, which is more than enough for the block-removal problem. However, we find that the open-source tabu algorithm~\cite{liang_libtsqubo} gives comparably good performance, reliably converging to the lowest-energy solutions in a matter of seconds for the hardest problems we can check. It can thus also be employed to further save resources even in the tractable case. We show a benchmark of tabu against exact solutions for up to the 29B-combination space in App.~\ref{sec:ed_tabu}.

    \subsection{\label{sec:exc}Candidate selection}

    \begin{figure}[t]
      \vskip 0.2in
      \begin{center}
        \centerline{\includegraphics[width=\columnwidth]{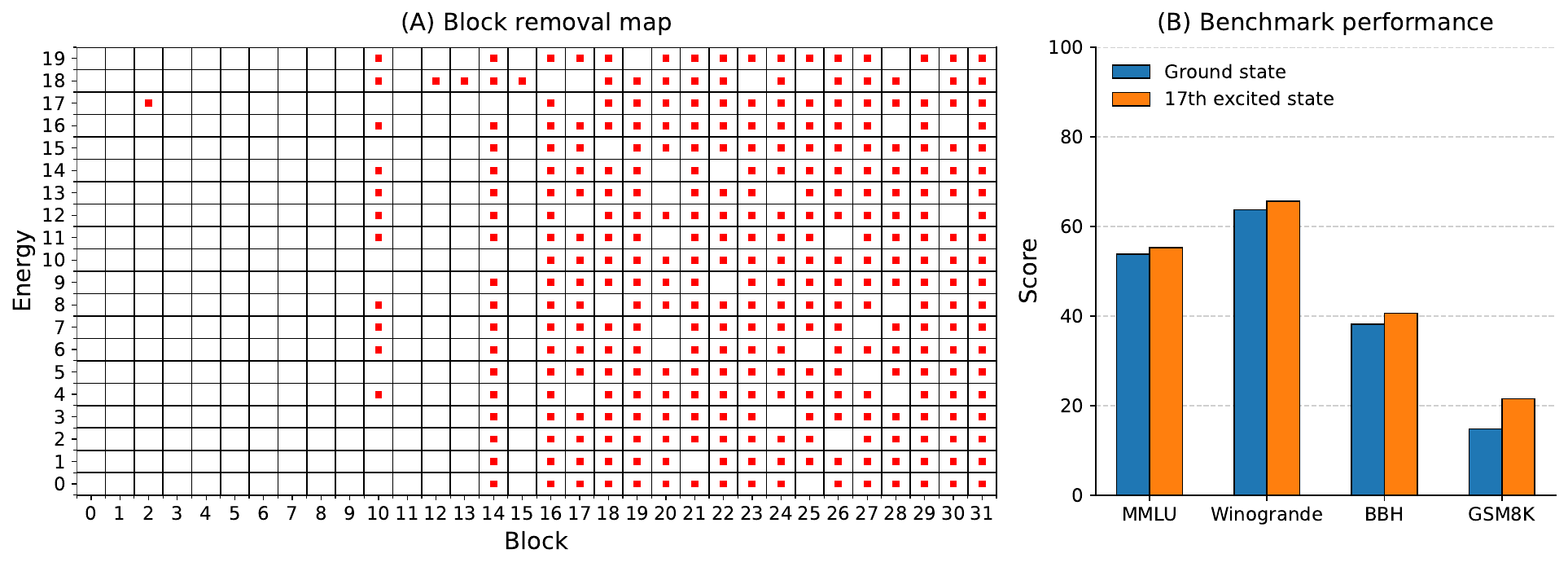}}
        \caption{
          (A) The number of times each block is recommended for removal among the first 20 low-energy states for Llama-3.1-8B-Instruct (16/32 blocks targeted for removal in total). Y-axis: state index, x-axis: transformer block index, red circle: block was recommended for removal in that state. (B) Selected benchmark results for states 0 (ground state) and 17.
        }
        \label{fig:llamafreq}
      \end{center}
    \end{figure}
  
    Our method naturally provides a set of multiple high-quality candidate configurations, and does not get stuck on a bad solution. Once Eq.~\eqref{eq:cbo} is set up, it is straightforward to examine both the ground state (lowest-energy configuration) and low-lying excited states. This allows alternative block-removal configurations to be explored without recomputing the Hessian.
    
    In Fig.~\ref{fig:llamafreq}, we show how often each block is selected for removal among the first 20 low-energy states  when removing 16/32 blocks for Llama-3.1-8B-Instruct. We see that the algorithm strongly favors removing blocks toward the end of the model, consistent with previous block-pruning approaches (see App.~\ref{app} for a list of block indices removed by different methods).
    In particular, the ground state suggests removing the blocks [14,16--24,26--31], and the first 16 low-energy states are quite similar. However, the 17th excited state, which removes the blocks [2,16,18--31], is the first configuration that proposes removing a block close to the beginning of the model, making it an interesting candidate for further inspection. 
    
    Notably, after short retraining, this excited-state configuration outperforms the ground state across several benchmarks (cf. Fig.~\ref{fig:llamafreq} (B)). This observation disproves the previous heuristic assumption that high-quality pruning solutions consist of consecutive blocks~\cite{gromov2025the} in the middle or towards the end of the model, and highlights the importance of addressing the full combinatorial nature of the problem. 

    It also shows that while the CBO energy is strongly correlated with downstream model performance, it is not perfectly correlated, and trying out several candidates above the ground state (though not too far from it in energy) is a crucial part of the process. In fact, CBO solvers are judged by their ability to reach the ground state, but we see that this is not even needed for our case. We rather need to quickly generate candidate solutions, which greatly relaxes the requirements on the solver.

 \subsection{\label{sec:noret}Results without retraining}

Table~\ref{table:benchmarks_unhealed} shows the full benchmark results without retraining. We see that CBO and BI are on par with each other and significantly outperform the other methods on MMLU for 8/32 removed blocks of Llama-3.1-8B-Instruct. For a deeper compression (16/32), all benchmarks become close to random, and retraining is needed for a meaningful comparison. For Qwen3-14B, our method is significantly better on MMLU (+10 percentage points for 12/40 removal).
For the larger Llama-3.3-70B-Instruct (Tab.~\ref{table:benchmarks_unhealed_Llama70B}), our method is again on par with BI for up to 24/80 removed blocks, but then significantly outperforms BI for the deep compression regime for 32/80 and 40/80 removed blocks, especially on MMLU (+23 percentage points). For 40/80, we note that CBO outperforms BI on all benchmarks, indicating that the consecutive solution found by BI, see App.~\ref{app}, is far from optimal.

\begin{table*}[t]
\centering
\caption{\label{table:benchmarks_unhealed}
Benchmarks without retraining. The \#br column gives the number of removed blocks, \#comb column gives the total number of all possible removal combinations $\binom{N}{M}$. The methods are: CBO:x = this paper, x stands for the state index, BI = block influence~\cite{men2025shortgpt}, SWM = Sliding-Window Merging~\cite{Ding_25}, Norm ratio~\cite{mistral3_26}.
}
\vspace{0.5\baselineskip}
\small
\setlength{\tabcolsep}{5pt}
\begin{tabular}{c r l !{\vrule width \arrayrulewidth} c !{\vrule width \arrayrulewidth} c c c c c}
\hhline{~~~|-|~~~~~}
\# br & \#comb & & MMLU & HellaSwag & Winogrande & ARC-C & BBH & GSM8K \\
\midrule

\multicolumn{9}{c}{\textbf{Llama-3.1-8B-Instruct} (32 blocks)} \\
\midrule
0  &             & Original & 68.3 & 59.8 & 73.7 & 53.6 &51.1 & 70.0 \\

\midrule
\multirow{4}{*}{8}
  & \multirow{4}{*}{10.5M}
  & CBO:0 & 63.1 & 40.9 & 63.6 & \textbf{39.1} & \textbf{46.8} & 0.0 \\
  %&  & BIit  & 0.55 & 0.42 & \textbf{0.67} & 0.35 & 0.41 & 0.00 \\
   &  & BI  & \textbf{65.4} & \textbf{43.2} & \textbf{67.2} & 36.9 & 45.5 & 0.0 \\
  &  & SWM & 24.1 & 31.1 & 59.5& 25.3 & 33.4 & 0.0 \\
  &  & Norm ratio & 26.0 & 41.9 & 56.8 & 28.4 & 29.6 & 0.0 \\

\midrule
\multirow{5}{*}{16}
  & \multirow{5}{*}{601M}
  & CBO:0 & 23.6 & 27.7 & 54.2 & 26.3 & 28.0 & 0.0 \\
  &  & CBO:17 & \textbf{28.7} & 29.1 & 53.5 & \textbf{29.8} & 29.5 & 0.0 \\
  %&  & BIit & 0.24 & \textbf{0.29} & 0.50 & 0.21 & 0.28 & 0.00 \\
    &  & BI  & 22.9 & 27.9 & \textbf{56.3}& 23.0 &26.6 & 0.0 \\
  &  & SWM & 23.4 & 28.2 & 50.0 & 22.2 & \textbf{30.1} & 0.0 \\
  &  & Norm ratio & 26.0 & \textbf{29.4} & 50.2 & 23.5 & 26.5 & 0.0 \\

\midrule
\multicolumn{9}{c}{\textbf{Qwen3-14B} (40 blocks)} \\
\midrule
0  &             & Original & 77.1 & 60.9 & 72.7 & 58.7 & 63.2 &89.2 \\

\midrule
\multirow{4}{*}{12}
  & \multirow{4}{*}{5.59B}
  & CBO:0 & \textbf{71.8} & 34.7 & \textbf{59.3} & \textbf{35.8} & \textbf{47.4} & 0.0 \\
  %&  & BIit & 0.50 & \textbf{0.41} & 0.58 & 0.32 & 0.42 & 0.00 \\
   &  & BI& 61.6 & \textbf{39.8} & 58.2 & 35.0 &43.6 & \textbf{0.1} \\
  &  & SWM & 50.5 & 36.4 & 54.5 & 29.1 & 34.4 & 0.0 \\
  &  & Norm ratio & 25.0 & 31.5 & 50.2 & 21.3 & 29.2 & 0.0 \\

\hhline{~~~|-|~~~~~}
\end{tabular}
\vspace{1em} 
\end{table*}
\begin{table*}[t]
\centering
\caption{\label{table:benchmarks_unhealed_Llama70B}
Benchmarks for the Llama-3.3-70B-Instruct model without retraining.
}
\vspace{0.5\baselineskip}
\small
\setlength{\tabcolsep}{5pt}
\begin{tabular}{c r l !{\vrule width \arrayrulewidth} c !{\vrule width \arrayrulewidth} c c c c c}
\hhline{~~~|-|~~~~~}
\# br & \#comb & & MMLU & HellaSwag & Winogrande & ARC-C & BBH & GSM8K \\
\midrule
\multicolumn{9}{c}{\textbf{Llama-3.3-70B-Instruct} (80 blocks)} \\
\midrule
0  &             & Original & 82.2 & 66.6 & 82.8 & 60.6 & 69.8 & 76.7 \\

\midrule
\multirow{3}{*}{8}
  & \multirow{3}{*}{29B}
  & CBO:0 & \textbf{81.8} & 64.1 &81.6 & 58.6 & \textbf{68.6} & 82.6 \\
  &  & BI & \textbf{81.8} & \textbf{64.3} & \textbf{81.9} & \textbf{59.1} & \textbf{68.6} & \textbf{90.1} \\
  &  & Norm ratio & 71.5 & 60.1 & 78.9 & 50.1 & 56.4 & 34.0 \\

\midrule
\multirow{3}{*}{16}
  & \multirow{3}{*}{$2.7\cdot10^{16}$}
  & CBO:0 & 80.7 & \textbf{61.4} & 79.3 & 53.0 & 61.3 & 60.6 \\
  &  & BI& \textbf{81.5} & 59.6 & \textbf{81.2} & \textbf{55.2} & \textbf{67.5} & \textbf{74.9} \\
  &  & Norm ratio & 43.0 & 51.3 & 64.7 & 34.8 & 37.5 & 4.2 \\

\midrule
\multirow{3}{*}{24}
  & \multirow{3}{*}{$1.6\cdot10^{20}$}
  & CBO:0 & 79.5 & 46.0 & 68.7 & 45.5 & 55.2 & 0.0\\
  &  & BI & \textbf{80.9} & \textbf{56.0} & \textbf{0.79} &\textbf{46.2} & \textbf{61.1} & \textbf{9.6} \\
  &  & Norm ratio & 35.4 &  46.8 & 65.7 & 34.4 & 35.9 & 4.3 \\
  \midrule
\multirow{3}{*}{32}
  & \multirow{3}{*}{$2.2\cdot10^{22}$}
  & CBO:0 & \textbf{76.6} & 39.0 & 68.5 & \textbf{39.8} & \textbf{51.4} & 0.0 \\
  &  & BI & 59.3& \textbf{48.0} & \textbf{74.5} & 36.5 & 53.4 & \textbf{1.4} \\
  &  & Norm ratio & 33.5 & 44.3 & 63.5 & 31.8 & 32.8 & 0.0 \\
  \midrule
\multirow{3}{*}{40}
  & \multirow{3}{*}{$10^{23}$}
  & CBO:0 & \textbf{76.9} & 34.9 & \textbf{69.0} & \textbf{37.5} & \textbf{48.8} & 0.0 \\
  &  & BI & 54.0 & 28.1 & 61.3 & 28.8& 34.7 & 0.0 \\
  &  & Norm ratio & 27.4 & \textbf{37.7} & 61.1 & 26.5 & 33.9 & 0.0 \\

\hhline{~~~|-|~~~~~}
\end{tabular}
\vspace{1em} 
\end{table*}

\subsection{\label{sec:ret}Results after retraining}
\begin{table*}[t]
\centering
\caption{\label{table:benchmarks_healed}
Benchmarks after retraining. Labels as in Tab.~\ref{table:benchmarks_unhealed}. Here, CBO:0 refers to the lowest energy state found by the tabu algorithm (the problem was only solved exactly for 8 blocks removed).
}
\vspace{0.5\baselineskip}
\small
\setlength{\tabcolsep}{5pt}
% \begin{tabular}{c r l c c c c c c}
\begin{tabular}{c r l !{\vrule width \arrayrulewidth} c !{\vrule width \arrayrulewidth} c c c c c}
\hhline{~~~|-|~~~~~}
\# br & \#comb & Model & MMLU & HellaSwag & Winogrande & ARC-C & BBH & GSM8K \\
\midrule

\multicolumn{9}{c}{\textbf{Llama-3.1-8B-Instruct} (32 blocks)} \\
\midrule
0  &             & Original & 68.3 & 59.8 & 73.7 & 53.6 &51.1 & 70.0 \\

\midrule
\multirow{4}{*}{8}
  & \multirow{4}{*}{10.5M}
  & CBO:0 &64.1 & 53.4 & 70.9 & 44.6 & 45.5 & 66.4 \\
  %&  & BIit & \textbf{0.65} & 0.52 & 0.71 & 0.44 & \textbf{0.48} & 0.61 \\
     &  & BI  & \textbf{65.4} & 52.9 & \textbf{71.6} & 44.7 &  \textbf{48.4} & \textbf{69.5} \\
  &  & SWM &58.1 & \textbf{53.6} & \textbf{71.6} & \textbf{45.4} & 44.2 & 63.4 \\
  &  & Norm ratio & 45.2 & 50.6 & 63.3 & 38.1 & 36.3 & 33.7 \\

\midrule
\multirow{5}{*}{16}
  & \multirow{5}{*}{601M}
  & CBO:0 & 53.8 & 41.9 & 63.7 & 32.8 & 38.1 & 14.8 \\
  &  & CBO:17 & 55.2 & \textbf{42.7} & 65.6 & \textbf{35.0} & 40.6 & 21.5 \\
  %&  & BIit & 0.43 & 0.39 & 0.60 & 0.30 & 0.35 & 0.14 \\
     &  & BI  & \textbf{56.9} & 41.8 & \textbf{66.5} & 32.2 & \textbf{42.1} & 10.1 \\
  &  & SWM & 49.1 & \textbf{42.7} & 63.2 & 30.6 & 40.1 & \textbf{33.2} \\
  &  & Norm ratio & 37.2 & 41.8 & 57.5 & 30.6 & 34.8 & 14.6 \\

\midrule
\multicolumn{9}{c}{\textbf{Qwen3-14B} (40 blocks)} \\
\midrule
0  &            & Original & 77.1 & 60.9 & 72.7 & 58.7 & 63.2 &89.2 \\

\midrule
\multirow{4}{*}{12}
  & \multirow{4}{*}{5.59B}
  & CBO:0 & \textbf{69.3} & 48.2 & 69.4 & 44.4 & \textbf{51.4} & 77.0 \\
   %&  & BIit  &0.63&\textbf{0.51} &0.71 & \textbf{0.47} & 0.47&0.75 \\
   &  & BI  & 66.0 & \textbf{51.0} & \textbf{72.0} & \textbf{47.4} & 48.8 & \textbf{78.6} \\
  &  & SWM & 61.4 & 49.1 & 68.9 & 44.5 & 45.1 & 72.1 \\
  &  & Norm ratio & 46.0 & 49.7 & 66.1 & 46.3 & 36.8 & 67.3 \\
\hhline{~~~|-|~~~~~}
\end{tabular}
\vspace{1em} 
\end{table*}
Table~\ref{table:benchmarks_healed} shows the benchmarks after retraining.
When removing 8/32 blocks from Llama-3.1-8B-Instruct, we see that BI marginally outperforms the CBO ground state (CBO:0). Notably, both still maintain significantly higher scores for MMLU than SWM and the Norm ratio. When we remove 16/32 blocks, we observe that the best-performing configuration (CBO:17) now outperforms BI on several benchmarks, and both retain a significant higher MMLU score than SWM and Norm ratio. For the 12/40 removal from Qwen3-14B, we see that CBO outperforms BI on MMLU and BBH, while BI scores higher on 4 of the other benchmarks.
%We interpret the strong MMLU performance as better retention of general knowledge, which is difficult to recover through retraining. In contrast, GSM8K scores are easier to improve with additional fine-tuning targeting reasoning style and output format. 
%We interpret the high scoring on MMLU as superior retention of general knowledge, which in our experience is hard to improve through further retraining. On the other hand, we want to note that GSM8K results should be interpreted with caution, as we empirically find that these scores are much easier to improve through additional fine-tuning (with some care devoted to targeting the output format and reasoning style).

    \subsection{Generalization to Heterogeneous and Mixture-of-Experts Architectures}

    % \begin{figure}[t]
    %   \vskip 0.2in
    %   \begin{center}
    %     \centerline{\includegraphics[width=0.8\columnwidth,height=5cm]{nemotron_benchmarks.pdf}}
    %     \caption{
    %       Benchmark results for NVIDIA-Nemotron-3-Nano-30B-A3B-FP8. We use the following notation: Gs:2 means the ground state (lowest-energy eigenvector) of the CBO problem when removing two layers, while, e.g., 19th excited:3 means the 19th excited state when removing 3 layers. See Tab.~\ref{tabel3} for the corresponding block index combinations.
    %     }
    %     \label{fig:nemotronbench}
    %   \end{center}
    % \end{figure}

    \begin{table}[t]
    \centering
    \caption{Benchmark results for NVIDIA-Nemotron-3-Nano-30B-A3B-FP8 when removing 2/52 and 3/52 blocks. CBO:x indicates the index in the low-energy spectrum.}
    \small
    \label{tab:nemotron_benchmarks}
    \begin{tabular}{c l c c c}
    \toprule
    \# br & Method & combination  & AIME25 &  GPQA  \\
    \midrule
    0 & Original & & 0.89 & 0.74\\
    \midrule
    \multicolumn{5}{c}{\textit{Removing 2 MoE layers}} \\
     \midrule
    \multirow{4}{*}{2}
      & CBO:0 & [8,10]           & 44.7 & 36.5 \\
      & CBO:1  & [8,45] &\textbf{86.3} & 67.2 \\
         & CBO:2  & [10,36] & 53.3 &45.7\\
      & CBO:10  & [8,38] & 63.3 & 63.6 \\
         & BI  & [38, 45] & 61.1 &\textbf{72.5}\\
      \midrule
    \multicolumn{5}{c}{\textit{Removing 3 MoE layers}} \\
     \midrule
    \multirow{2}{*}{3}
      & CBO:0  & [8,10,38]         & 28.3 & 33.4 \\
       & CBO:1 & [8,10,40] & 39.2 & 32.9 \\
      & CBO:22 & [8,38,45] & \textbf{62.0} & \textbf{65.1} \\
       & BI & [38,40,45] & 53.9 & 64.1 \\
        \midrule
    \multicolumn{5}{c}{\textit{Removing 2 attention layers}} \\
     \midrule
         \multirow{4}{*}{2}
      & CBO:0 & [5, 33]           & 0.0 & 29.0 \\
      & CBO:1  & [5,12] & 28.7 & 43.6 \\
        & CBO:7  & [42,5] & \textbf{80.7} & \textbf{68.1} \\
            & BI  & [33,42] & 0.0 & 23.9 \\
    \bottomrule
    \end{tabular}
    \vspace{1em}
    \end{table}

    \begin{table}[t]
        \centering
        \caption{
        Evaluation for NVIDIA-Nemotron-3-Nano-30B-A3B-FP8 on a broader suite of six benchmarks for selected configurations from Table~\ref{tab:nemotron_benchmarks}. Mean denotes the unweighted average across tasks. CBO:x indicates the index in the low-energy spectrum. Note that AIME25 and GPQA scores may differ slightly from Table~\ref{tab:nemotron_benchmarks} due to the probabilistic nature of LLM inference and different sampling parameters across evaluation runs.}
        \label{tab:sixbench}
        \small
        \begin{adjustbox}{max width=\textwidth}
        \begin{tabular}{c l c c c c c c c c}
        \toprule
        \# br & Method & Combination & AIME25 & GPQA & MMLU-Pro & SciCode & IFBench & LCB & Mean \\
        \midrule
        0 & Original & & 89.3 & 73.9 & 78.9 & 32.5 & 72.9 & 74.9 & 70.4 \\
        \midrule
        \multicolumn{10}{c}{\textit{Removing 2 MoE layers}} \\
        \midrule
        \multirow{2}{*}{2}
          & CBO:1  & [8,45]  & \textbf{85.0} & 69.5& 74.8 & 27.0 & 69.5 & \textbf{71.2} & \textbf{66.2} \\
          & BI     & [38,45] & 65.0 & \textbf{70.1} & \textbf{76.4} & \textbf{28.5} & \textbf{70.7} & 68.4 & 63.2 \\
        \midrule
        \multicolumn{10}{c}{\textit{Removing 3 MoE layers}} \\
        \midrule
        \multirow{3}{*}{3}
          & CBO:22 & [8,38,45]  & \textbf{60.3} & \textbf{64.4} & \textbf{73.3} & 8.7 & \textbf{70.1} & \textbf{65.9} & \textbf{57.1} \\
          & CBO:1  & [8,10,40]  & 37.0 & 35.2 & 52.2 & 10.6 & 54.4 & 30.05 & 36.6 \\
          & CBO:0  & [8,10,38]  & 32.7 & 36.0 & 52.7 & \textbf{12.0} & 53.0 & 35.1 & 36.9 \\
        \bottomrule
        \end{tabular}
        \end{adjustbox}
        \vspace{1em}
        \end{table}
        
    Finally, we evaluate our method on NVIDIA-Nemotron-3-Nano-30B-A3B-FP8~\cite{nvidia_nemotron_nano_v3_2025}, a heterogeneous hybrid transformer architecture significantly more challenging than standard dense transformers. The model interleaves Mamba2 layers, attention layers, and MoE layers in a non-uniform block structure, making block-removal decisions substantially more complex. The layout of the model can be summarized as
    \textit{MEMEM*EMEMEM*EMEMEM*EMEMEM*
EMEMEM*EMEMEMEM*EMEMEMEME},
    where \textit{M} denotes a Mamba2 layer, \textit{*} an attention layer, and \textit{E} an MoE layer (52 layers in total).
    
    % Benchmarks are evaluated on GPQA~\cite{gpqa_23} and AIME25 using the Nemo Skills library~\cite{nemo_skills}. As calibration data for the Hessian, we use 2048 samples of synthetic data covering code, mathematics, science, chat, and agentic tasks.

    We study whether blocks can be removed at all from this complex architecture without retraining, while preserving downstream performance. We now use a custom calibration dataset spanning mathematics, coding, scientific reasoning, and conversational tasks and focus on solutions of the CBO where either 2-3 MoE or 2 attention layers are removed. The former case corresponds to  $\sim8\%$ and $\sim12\%$ parameter reduction, while the latter case improves inference speed.
    Evaluation is performed with the Nemo-Skills framework~\cite{nemo_skills} on the full AIME25 and GPQA~\cite{gpqa_23} benchmarks, using repeated sampling (10 rollouts for AIME25 and 5 for GPQA) to obtain stable estimates.
    % \textcolor{red}{Ali: add evaluation details}
    
    Table~\ref{tab:nemotron_benchmarks} shows the benchmark results. We observe that the first excited state for the 2/52 case and the 22nd excited state for the 3/52 case still maintain strong scores. Across all configurations, AIME25 is relatively sensitive, indicating that mathematical reasoning relies more critically on specific expert layers, whereas GPQA exhibits greater robustness. Attention layer removal is generally less stable and often leads to severe degradation, although isolated configurations such as [42,5] remain viable. 

    To validate that these trends generalize beyond the two primary benchmarks, Table~\ref{tab:sixbench} reports results on a broader suite of six benchmarks. The best-performing 2-layer configuration ([8,45]) retains strong overall performance (0.66 vs.\ 0.70 mean accuracy for the baseline), confirming that capability preservation extends beyond AIME25 and GPQA. Note that AIME25 and GPQA scores may differ slightly from Table 4 due to the probabilistic nature of LLM inference and different sampling parameters across evaluation runs. Overall, these results demonstrate that structured redundancy in large hybrid models is unevenly distributed and can be effectively exploited via careful selection, enabling meaningful compression without retraining, while preserving strong reasoning performance.

\subsection{Limitations} \label{sec:limits}

The algorithm presented here comes with some limitations that open future research opportunities. In App.~\ref{sec:dataset}, we demonstrate that the selection of a high-quality dataset matters, and it might be possible to find an even stronger datamix that further improves the results. We also observe that the ground state does not always yield the best configuration for all downstream tasks and had to select the candidate solutions manually, but an automated selection procedure would be desirable. The CBO problem is challenging to solve at larger scales, and while we have found that the tabu solver works quite well, it would be valuable to explore a broader benchmarking of CBO solvers. Lastly, we see that even though our algorithm often achieves better scores for MMLU, it may perform worse on other benchmarks. Thus, it seems that there is no one-size-fits-all block removal algorithm, demonstrating the need for tailored approaches depending on the problem. Gaining a better understanding of when and how each algorithm performs well is an important question for practical future applications.

\section{Conclusion} \label{sec:concl}

    In this work, we have introduced an algorithm based on CBO (constrained binary optimization) for selecting which combination of transformer blocks to remove from LLMs. Our approach introduces an auxiliary variable for each block (0: keep, 1: remove) and uses a second-order Taylor expansion of the loss to explicitly take into account interactions between blocks. The energies of the resulting CBO problem strongly correlate with high performance across multiple benchmarks. Importantly, we find that the whole low-energy spectrum contains many high-quality solutions, allowing practitioners to select configurations that best match specific accuracy or structural requirements.
    
    We have applied our method to Llama-3.1-8B-Instruct, Qwen3-14B and Llama-3.3-70B-Instruct and have shown that the resulting pruned models are on par or significantly outperform those obtained by state-of-the-art block-removal algorithms across a range of benchmarks. Most notably, our models consistently achieve substantially higher MMLU scores, indicating superior retention of general knowledge, in particular for deep compression.
    
    We have further demonstrated that our approach generalizes beyond homogeneous dense transformers to hybrid architectures with MoE layers. In particular, for NVIDIA-Nemotron-3-Nano-30B-A3B-FP8, we successfully removed 2-3 MoE and 2 attention layers while maintaining high accuracy on GPQA and AIME25, even without any retraining. This highlights the robustness of our formulation in challenging, non-homogeneous architectural settings.
    
Efficiently compressing LLMs while preserving their knowledge remains an active and rapidly evolving research area, and several promising extensions of our work remain. For example, one could further study the influence of the calibration dataset or analyze the correlation between solver quality and downstream tasks. Finally, we note that our method is compatible with layer-merging approaches~\cite{Ding_25}, quantization~\cite{frantar2022gptq,lin2023awq}, low-rank approximation~\cite{yuan_23,wang2025svdllm,rausch_25}, width pruning~\cite{sun2023wanda,muralidharan_24,guo2025slimllm,Gili_25}, and weight clustering~\cite{aizpurua2026}. Exploring how to effectively combine these techniques is left for future work.

   %Efficiently compressing LLMs while preserving their knowledge remains an active and rapidly evolving research area, and several promising extensions of our work remain. For example, one could further study the influence of the calibration dataset to determine if one can make specialized models by e.g. only using data from certain knowledge domains. It would also be interesting to analyze the correlation between solver quality and downstream tasks in cases where exact solutions are not available. Finally, we note that our method is compatible with layer-merging approaches~\cite{Ding_25}, quantization~\cite{frantar2022gptq,lin2023awq}, low-rank approximation~\cite{yuan_23,wang2025svdllm,rausch_25}, width pruning~\cite{sun2023wanda,muralidharan_24,guo2025slimllm,Gili_25}, and weight clustering~\cite{aizpurua2026}. Exploring how to effectively combine these techniques is left for future work.

Our data and code are available at~\cite{rausch_2026_20125262,jansen2026block_code}.
\section*{Acknowledgements}
 We acknowledge Donostia International Physics Center (DIPC), Ikerbasque, Basque
Government, Diputaci\'on de Gipuzkoa, European Innovation Council (EIC), Institute for Advanced Studies in Basic Sciences (IASBS), and Spanish Government for constant support as well as insightful discussions with Angus Dunnett, Sukhbinder Singh, Safa Hamreras, and Alejo Lopez Avila.
\bibliography{references}
\bibliographystyle{plainnat}

%%%%%%%%%%%%%%%%%%%%%%%%%%%%%%%%%%%%%%%%%%%%%%%%%%%%%%%%%%%%

\appendix

\section{\label{app}Block combinations}

In Table \ref{tabel3}, we show which blocks were removed by the different methods. For SWM, we show which layers were merged together.
\begin{table*}[t]

\centering
\caption{Blocks removed by different models and methods.}
\vspace{0.5\baselineskip}
\label{tab:blocks_removed}

\small
\setlength{\tabcolsep}{5pt}

\begin{tabular}{c l l p{7.5cm}}
\toprule
\# br & Method & Blocks removed \\
\midrule

\multicolumn{3}{c}{\textbf{Llama-3.1-8B-Instruct} (32 blocks)} \\
\midrule
8  & CBO:0 & [20, 23, 25, 27, 28, 29, 30, 31] \\
8  & BI & [21--28] \\
8  & SWM & [[6,7], [12,13], [24--30]] \\
8  & Norm ratio & [8, 10, 11, 12, 13, 14, 19, 20] \\

\midrule
16  & CBO:0 & [14, 16, 17, 18, 19, 20, 21, 22, 23, 24, 26, 27, 28, 29, 30, 31] \\
16  & CBO:17 & [2, 16, 18, 19, 20, 21, 22, 23, 24, 25, 26, 27, 28, 29, 30, 31] \\
16  & BI & [15--30] \\
16  & SWM & [[6,7], [8--10], [14--16], [19--30]] \\
16  & Norm ratio & [8, 9, 10, 11, 12, 13, 14, 15, 19, 20, 22, 23, 24, 25, 27, 29] \\

\midrule
\multicolumn{3}{c}{\textbf{Qwen3-14B} (40 blocks)} \\
%\midrule
%8  & CBO:0 & [30, 32, 33, 35, 36, 37, 38, 39] \\
%8  & BI & [35, 34, 33, 32, 22, 2, 2, 28] \\
%8  & SWM & [[18,19], [23--25], [32--34], [35--38]] \\
%8  & Norm ratio & [11, 12, 13, 14, 15, 16, 18, 39] \\

\midrule
12  & CBO:0 & [2, 26, 30, 31, 32, 33, 34, 35, 36, 37, 38, 39] \\
12  & BI& [2, 3, 22--24, 31--37] \\
12  & SWM& [[14--16], [21,22], [26,27], [24,25], [31--38]] \\
12  & Norm ratio & [10, 11, 12, 13, 14, 15, 16, 17, 18, 20, 23, 39] \\

\midrule
\multicolumn{3}{c}{\textbf{Llama-3.3-70B-Instruct} (80 blocks)} \\
\midrule
8  & CBO:0 & [46, 56, 58, 59, 62, 64, 65, 67] \\
8  & BI & [54, 58,59,61--63,65,66] \\
\midrule
16 & CBO:0 & [46, 55--62, 64--66, 68, 71, 72, 74] \\
16 & BI & [50, 53--67] \\
\midrule
24 & CBO:0 & [43, 49, 50, 53--56, 58--61, 63-66, 68--72, 76--79]\\
24 & BI & [46--51, 53--69, 71]\\
\midrule
32 & CBO:0 & [40, 41, 43, 47, 49--54, 56--61, 63--66, 68--79] \\
32 & BI & [42--73] \\
\midrule
40 & CBO:0 & [29, 40, 41, 43, 44--79] \\
40 & BI& [37--76] \\
\bottomrule
\end{tabular}
\label{tabel3}
\vspace{1em} 
\end{table*}

\section{\label{sec:ed_tabu} Comparison of exact and approximate CBO solutions}

\begin{figure}
  \centering
  \includegraphics[width=\textwidth]{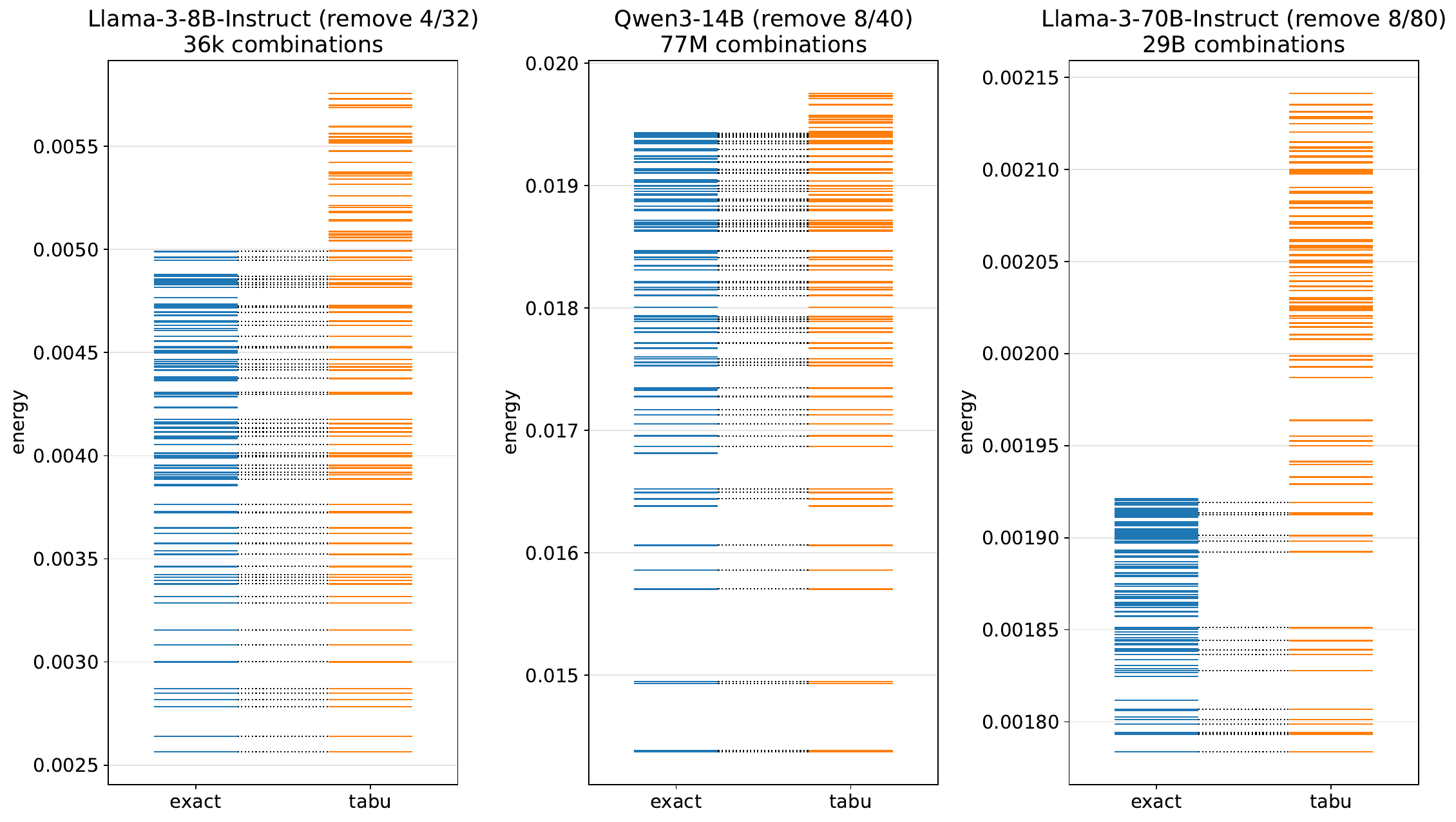}
  \caption{
  \label{fig:exact_tabu}
  Comparison between the exact low-energy spectrum and the solutions obtained by the tabu algorithm for an easy, medium and hard CBO problem. Tabu was set to provide 1000 solutions, we show the lowest 100 in both cases for clarity. Matching solutions are joined by dotted lines.}
\end{figure}

Figure~\ref{fig:exact_tabu} compares the exact low-energy states with those obtained by the tabu algorithm. We find that tabu reliably outputs the ground state for all cases shown, including the space of $\binom{80}{8}\approx29\text{B}$ combinations for Llama-3.3-70B-Instruct. It is also able to output the lowest excited states, but fails to fully reproduce the spectrum further up, especially for harder problems.
The running time is only seconds, which completely fulfils our requirement of quickly generating several candidate solutions. A faithful reconstruction of the full low-energy spectrum is not a requirement for our problem.
However, we note that we can in principle run several solvers in parallel or further iterate over tabu given the cheap running time, in order to generate even more solutions.

\section{\label{sec:grad}Influence of the first-order term}

We include the first-order (gradient) term with $\beta=1$ in Eq.~\eqref{eq:taylorexp} for Qwen3-14B and compare with the case of removing 12/40 blocks. Table~\ref{tab:grad} shows that this renders the benchmarks random even before retraining. Thus, we set $\beta=0$ in our other experiments and do not further pursue the tuning of $\beta$ as a hyperparameter.

Our tentative explanation of why this is happening is that the second-order term is good proxy for the interaction between blocks (it captures the effect of the influence of the removal of one block upon another block), while the addition of the gradient puts too much emphasis on a local score and washes out this interaction effect.

\begin{table}[t]
\centering
\caption{Effect of retaining the first-order loss term.}
\label{tab:grad}
\small
\begin{tabular}{c l c c c c c c}
\toprule
\# br & Method & MMLU & HellaSwag & Winogrande & ARC-C & BBH & GSM8K \\
\midrule
\multicolumn{8}{c}{\textbf{Qwen3-14B} (40 blocks)} \\
\midrule
12 & CBO:0             & \textbf{71.8} & 34.7 & \textbf{59.3} & \textbf{35.8} & \textbf{47.4} & \textbf{0.0} \\
12 & CBO + 1st order: 0 & 25.4          & \textbf{36.3}          & 50.1          & 25.2          & 31.3          & \textbf{0.0}          \\
\bottomrule
\end{tabular}
\vspace{1em}
\end{table}

\section{\label{sec:dataset}Influence of the dataset}
Here, we analyze the influence of different calibration datasets on the Qwen3-14B model. We report results for OpenHermes-2.5~\cite{OpenHermes25} (used in the main text) and compare them with those obtained using wikitext2~\cite{wikitext} and LMSYS-Chat-1M~\cite{lmsys}, both for the ground state (0) and 1st excited state (1), using 2048 samples from each dataset. Overall, we observe that calibration datasets can be broadly categorized into low- and high-quality groups. While low-quality datasets should be avoided, OpenHermes-2.5 and LMSYS-Chat-1M clearly fall into the high-quality category.
\begin{table}[ht]
\centering
\caption{Benchmark results using different calibration datasets for Qwen3-14B.}
\begin{tabular}{c l l !{\vrule width \arrayrulewidth} c !{\vrule width \arrayrulewidth} c c c c c}
\hhline{~~~|-|~~~~~}
\# br & Dataset & Method & MMLU & Hellaswag & Winogrande & ARC & BBH & GSM8K \\
\midrule
\multicolumn{9}{c}{\textbf{Qwen3-14B} (40 blocks)} \\
\midrule
%8  & OpenHermes-2.5 & CBO:0 & 0.70 & 0.38 & 0.61 & 0.39 & 0.51 & 0.0 \\
%   & lmsys-chat-1m  & CBO:0 & \textbf{0.73} & 0.37 & 0.61 & \textbf{0.40} & \textbf{0.52} & 0.0 \\
%   & lmsys-chat-1m  & CBO:1 & 0.30 & 0.33 & 0.51 & 0.22 & 0.39 & 0.0 \\
%   & Wikitext       & CBO:0 & 0.51 & 0.46 & \textbf{0.62} & 0.37 & 0.33 & 0.0 \\
%   & Wikitext       & CBO:1 & 0.28 & \textbf{0.47} & 0.60 & 0.39 & 0.34 & 0.0 \\
%\midrule
12 & OpenHermes-2.5 & CBO:0 & \textbf{71.8} & 34.7 & 59.3 & 35.8 & 47.4 & 0.0 \\
   & lmsys-chat-1m  & CBO:0 & 63.0 & 34.1 & \textbf{61.2} & 34.7 & \textbf{48.5} & 0.0 \\
   & lmsys-chat-1m  & CBO:1 & 60.7 & 34.8 & 58.2 & \textbf{36.7} & 46.9 & 0.0 \\
   & Wikitext       & CBO:0 & 27.1 & 32.4 & 54.5 & 28.7 & 30.1 & 0.0 \\
      & Wikitext       & CBO:1 & 31.9 & 36.3 & 54.8 & 30.6 & 33.4& \textbf{0.1} \\
\hhline{~~~|-|~~~~~}
\end{tabular}
\vspace{1em}
\label{tab:benchmark-results}
\end{table}

\section{\label{sec:statistical-significance}Statistical Significance} 
Table~\ref{tab:nemotron_stderr} reports AIME25 and GPQA performance together with 
the standard error of the mean (SEM), computed over multiple stochastic rollouts 
per instance. Specifically, we use 10 rollouts per problem for AIME25 and 5 rollouts per question for GPQA, yielding 300 and 990 total samples, respectively.

Across all evaluated configurations, the standard error of the mean (SEM) remains consistently low (typically $\leq 0.02$ absolute), indicating that the reported performance estimates  are stable with respect to decoding stochasticity. The only exception is CBO $[45,38,8]$ on AIME25, where the SEM slightly increases to $0.03$, reflecting higher sensitivity under aggressive compression. Importantly, the performance differences between methods are substantially larger than the observed uncertainty. For instance, the gap between CBO $[45,8]$ and BI $[45,38]$ reaches 20 percentage points on AIME25, while the corresponding SEM is two orders of magnitude smaller. This suggests that the observed improvements are statistically meaningful despite the absence of repeated full training runs.
We therefore focus our analysis on configurations where the effect size clearly  exceeds the estimated variability and avoid over-interpreting small differences within the noise level.
    
\begin{table}[tbh]
\centering
\caption{
AIME25 and GPQA accuracy (in \%) with standard error of the mean (SEM) for 
\textsc{Nemotron-3-Nano-30B-A3B}. Results are computed over multiple stochastic rollouts 
(10 for AIME25, 5 for GPQA).}
\label{tab:nemotron_stderr}
\small
\setlength{\tabcolsep}{6pt}
\renewcommand{\arraystretch}{1.05}
\begin{tabular}{lcc}
\toprule
\textbf{Model} & \textbf{AIME25} & \textbf{GPQA} \\
\midrule
Baseline & $89.33 \pm 0.01$ & $73.94 \pm 0.01$ \\
\midrule
\multicolumn{3}{l}{\textit{Block Influence (BI)}} \\
BI [45,38] & $65.00 \pm 0.01$ & $\mathbf{70.10 \pm 0.01}$ \\
BI [42,33] & $0.00 \pm 0.00$  & $23.94 \pm 0.01$ \\
\midrule
\multicolumn{3}{l}{\textit{Constrained Binary Optimization (CBO)}} \\
CBO [45,8] & $\mathbf{85.00 \pm 0.01}$ & $69.49 \pm 0.01$ \\
CBO [42,5] & $80.67 \pm 0.01$ & $68.08 \pm 0.00$ \\
CBO [38,8] & $63.33 \pm 0.02$ & $63.64 \pm 0.00$ \\
CBO [45,38,8] & $60.33 \pm 0.03$ & $64.34 \pm 0.01$ \\
CBO [40,10,8] & $37.00 \pm 0.02$ & $35.15 \pm 0.01$ \\
CBO [38,10,8] & $32.67 \pm 0.02$ & $35.96 \pm 0.01$ \\
CBO [10,8] & $46.67 \pm 0.02$ & $38.89 \pm 0.01$ \\
CBO [26,5] & $42.00 \pm 0.02$ & $47.78 \pm 0.01$ \\
CBO [12,5] & $28.67 \pm 0.02$ & $43.64 \pm 0.01$ \\
CBO [33,5] & $0.00 \pm 0.00$  & $28.99 \pm 0.01$ \\
\bottomrule
\end{tabular}
\end{table}

\section{\label{sec:broader}Broader impacts}
The goal of this work is to advance the development of effective compression algorithms for LLMs. Progress in this area can have a meaningful societal impact by reducing the computational and hardware requirements needed to deploy and fine-tune LLMs, thereby lowering energy consumption and enabling broader access to these technologies. By facilitating more efficient deployment on resource-constrained systems, such methods can contribute to the responsible and more equitable use of LLMs.

\section{\label{ret}Hyper parameters for the retraining}
Here we show the hyper parameters used for the retraining of the Llama-3.1-8B-Instruct and Qwen3-14B models.
\begin{table}[t]
\centering
\caption{Training hyperparameters used for retraining the compressed Llama-3.1-8B-Instruct and Qwen3-14B models. The only difference was that Per-device train/eval batch sizes were 4 for Qwen3-14B.}
\label{tab:training_hparams}
\begin{tabular}{ll}
\toprule
\textbf{Hyperparameter} & \textbf{Value} \\
\midrule
Model dtype & bfloat16 \\
Teacher model dtype & bfloat16 \\
Dataset & OpenHermes-2.5 \\
Validation split & 0.01 \\
Random seed & 42 \\
Dataset shuffling & True \\
\midrule
Max training steps & 10\,000 \\
Training epochs & 1 \\
Per-device train batch size & 16 \\
Per-device eval batch size & 16 \\
Gradient accumulation steps & 1 \\
Learning rate & $2.0 \times 10^{-5}$ \\
Weight decay & $1.0 \times 10^{-4}$ \\
Warmup steps & 100 \\
Gradient clipping & 1.0 \\
Learning-rate scheduler & Cosine \\
Maximum sequence length & 1024 \\
\bottomrule
\end{tabular}
\end{table}
%%%%%%%%%%%%%%%%%%%%%%%%%%%%%%%%%%%%%%%%%%%%%%%%%%%%%%%%%%%%

\clearpage

\newpage
%v\input{checklist.tex}

\end{document}